%% file: main.tex
\documentclass[10pt,twocolumn,letterpaper]{article}

\usepackage{cvpr}              


\definecolor{cvprblue}{rgb}{0.21,0.49,0.74}
\usepackage[pagebackref,breaklinks,colorlinks,allcolors=cvprblue]{hyperref}
\usepackage{graphicx}
\usepackage{capt-of}
\usepackage{afterpage}
\usepackage{float}
\usepackage{amsmath}
\usepackage{hyperref}
\usepackage{multirow}
\usepackage{caption}

\def\paperID{11036} 
\def\confName{CVPR}
\def\confYear{2025}

\title{Zero-Shot Head Swapping in Real-World Scenarios}

\author{Taewoong Kang$^{1}$\thanks{Equal contribution} \quad Sohyun Jeong$^{1,2}$\footnotemark[1] \quad Hyojin Jang$^{1}$\footnotemark[1] \quad Jaegul Choo$^{1,2}$ \\
$^{1}$KAIST $^{2}$ FLIPTION \\
\texttt{\{keh0t0, jsh0212, wkdgywlsrud, jchoo\}@kaist.ac.kr} }

\begin{document}
\begin{figure}
\twocolumn[{
\renewcommand
\twocolumn[1][]{#1}
\centering 
\maketitle
\includegraphics[width=1.\linewidth]{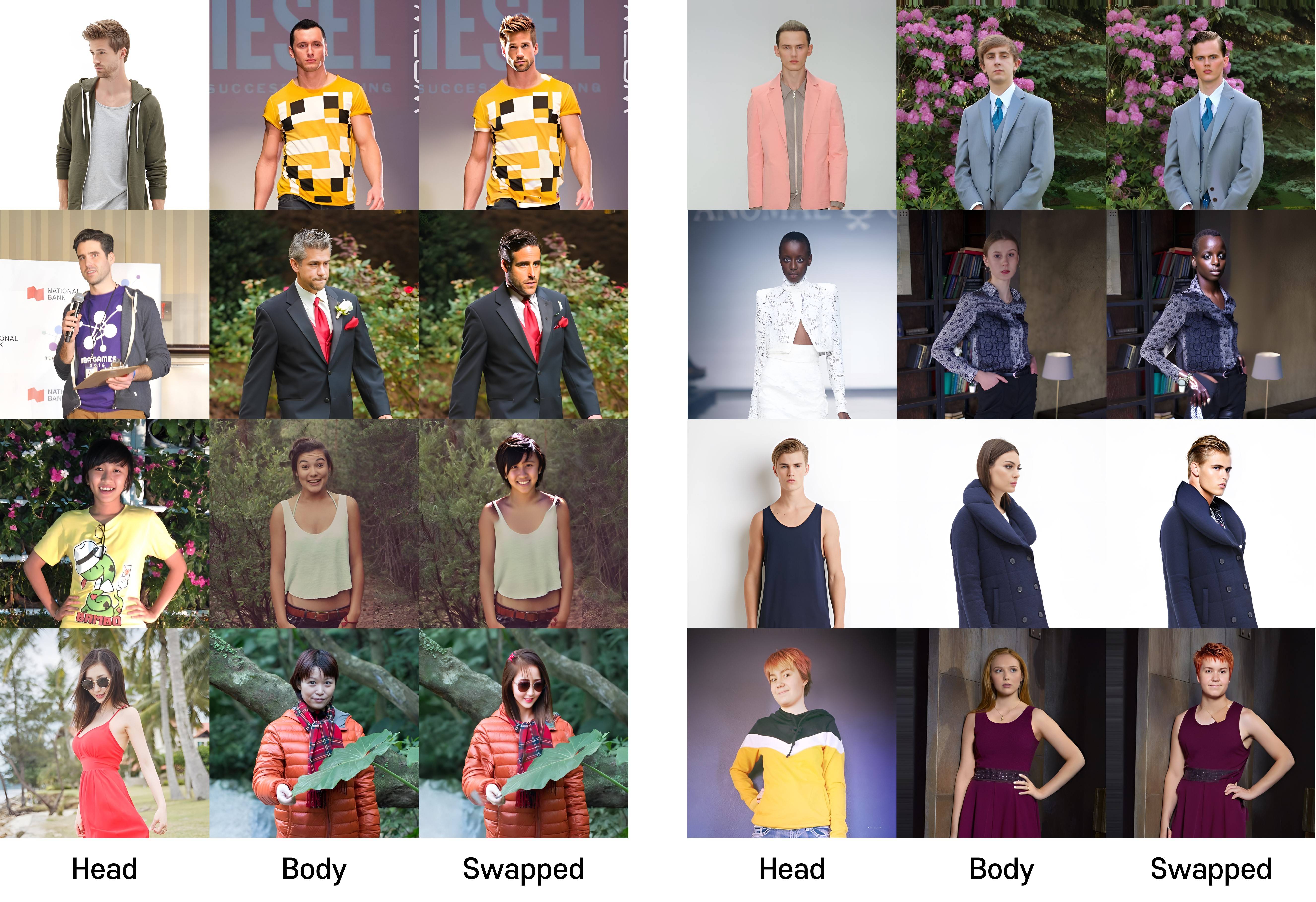}
\caption{\textbf{Head-swapped images generated by our approach.} 
Using the proposed method, HID, the head in the images of \textit{Head} column is seamlessly integrated onto the images of \textit{Body} column, resulting in realistic and cohesive head-swapped images in the \textit{Swapped} column.}
\label{fig:teaser}
\vspace{\baselineskip}
}]
\end{figure}

\input{sec/0_abstract}    
\input{sec/1_intro}
\input{sec/2_relatedwork}
\input{sec/3_method}

\input{sec/4_experiments}
\input{sec/5_conclusion}
\input{sec/Acknowledgment}
{
    \small
    \bibliographystyle{ieeenat_fullname}
    \bibliography{main}
}
\input{sec/X_suppl}

\end{document}

%% file: sec/0_abstract.tex
\begin{abstract}

With growing demand in media and social networks for personalized images, the need for advanced head-swapping techniques—integrating an entire head from the head image with the body from the body image—has increased. However, traditional head-swapping methods heavily rely on face-centered cropped data with primarily frontal-facing views, which limits their effectiveness in real-world applications. 
Additionally, their masking methods, designed to indicate regions requiring editing, are optimized for these types of dataset but struggle to achieve seamless blending in complex situations, such as when the original data includes features like long hair extending beyond the masked area. To overcome these limitations and enhance adaptability in diverse and complex scenarios, we propose a novel head swapping method, \textbf{HID}, that is robust to images including the full head and the upper body, and handles from frontal to side views, while automatically generating context-aware masks. For automatic mask generation, we introduce the IOMask, which enables seamless blending of the head and body, effectively addressing integration challenges. We further introduce the hair injection module to capture hair details with greater precision. Our experiments demonstrate that the proposed approach achieves state-of-the-art performance in head swapping, providing visually consistent and realistic results across a wide range of challenging conditions.

\end{abstract}

%% file: sec/1_intro.tex
\section{Introduction}
\label{sec:intro}

With recent advancements in media and social networks, the demand for face-editing has increased, enabling users to replace faces with those of celebrities or fictional characters. To this end, continuous developments has been made in Face Swapping~\cite{chen2020simswap, li2019faceshifter, deepfacelab, nirkin2019fsgan, zhu2021one}. However, face swapping only replaces the face identity (face ID) that included eyes, nose, lip, eyebrow and skin, which limits its realism in applications like virtual avatar generation, movie and advertisement synthesis, and social media content creation, where differences in facial shape and hairstyle can make the results less convincing. In these cases, head swapping becomes essential.

\begin{figure}[t]
  \centering
   \includegraphics[width=\linewidth]{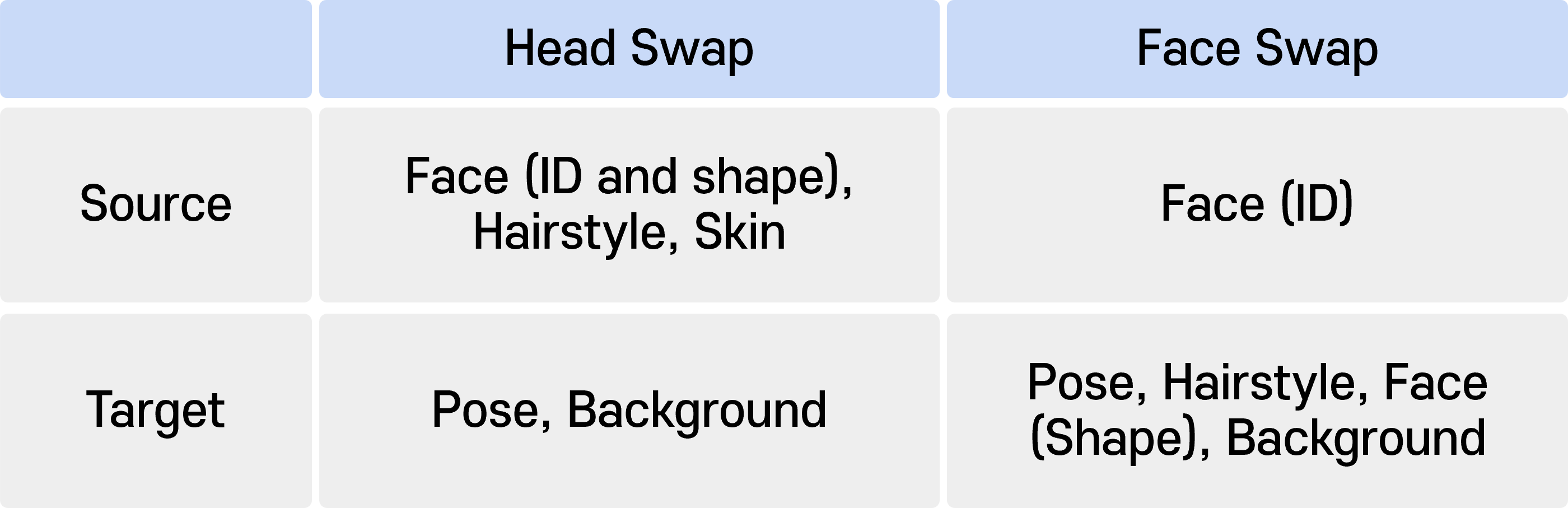}
   \caption{Face swapping simply applies the head image’s face ID to the body image. In contrast, head swapping requires applying not only the head image’s face ID but also the hairstyle, face shape, and skin tone.
}
   \vspace{-\baselineskip}
   \label{fig:headswapDef}
\end{figure}

Unlike face swapping, head swapping combines the entire head, including the face ID, face shape, and hairstyle, of the head image with the body image's body, as their differences are highlighted in \cref{fig:headswapDef}. Therefore, while face swapping only applies the face ID to the body image, head swapping needs to seamlessly merge the entire head from the head image with the body in the body image. Due to its increased complexity, head swapping still faces numerous unresolved challenges.
1) For a realistic head swap, it requires seamless blending between the swapped head and the original body. 
2) Unlike face swapping, where only the face ID features of the head image are extracted, head swapping have to also handle the structural information of the head that includes a hairstyle and a facial structure. 

\begin{figure}[t]
  \centering
   \includegraphics[width=\linewidth]{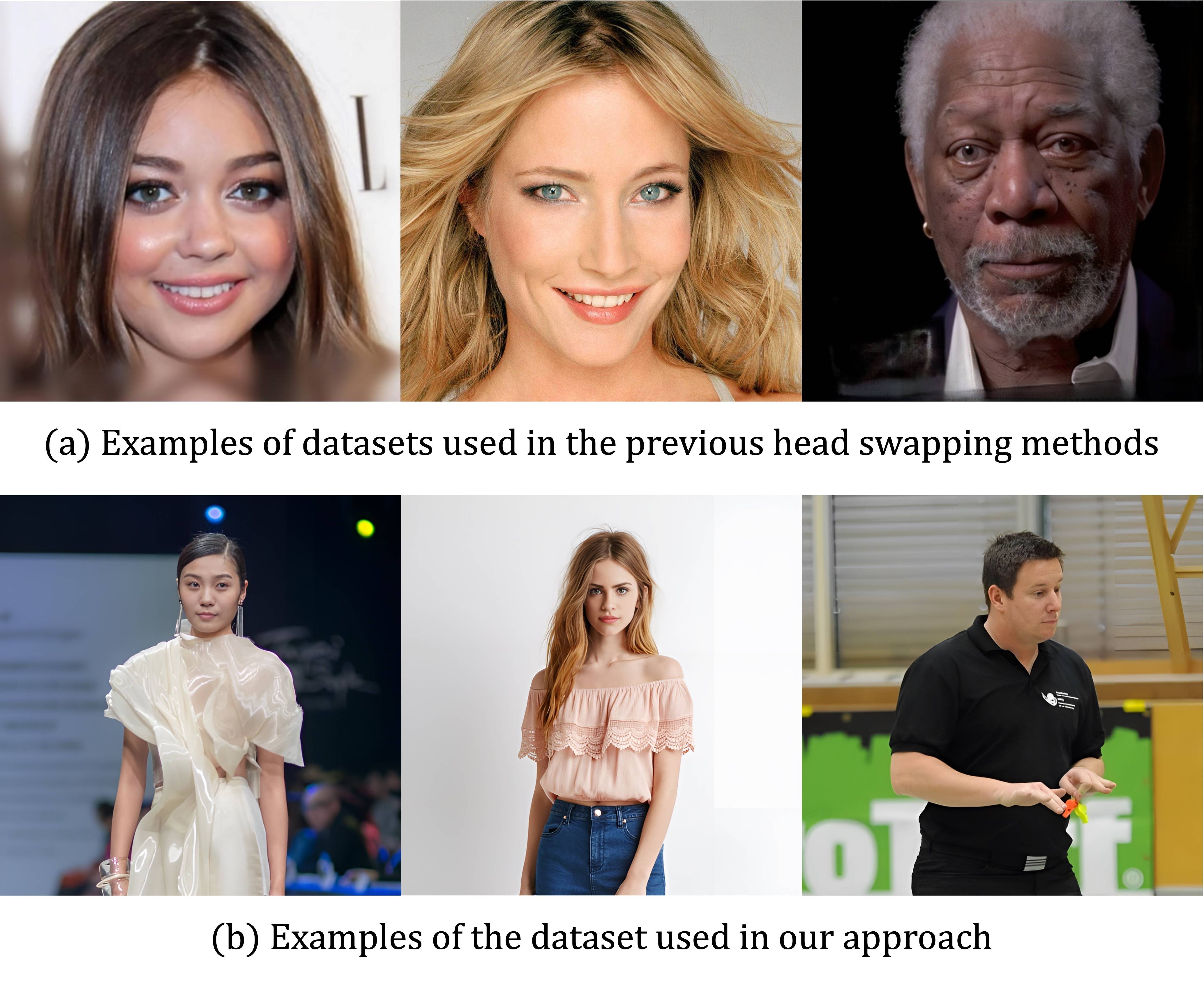}
   \caption{
   Most previous head-swapping methods~\cite{deepfacelab, reface, faceX} rely on datasets that require involving the zoomed-in, face-centered images as shown in (a). In this case, an additional step is needed to merge the head-swapped images back onto the original body, which can often result in mismatched or unnatural outcomes. In contrast, our approach uses the dataset that includes the whole upper body, as shown in (b), enabling head swapping in more realistic, real-world scenarios.
}
   \vspace{-\baselineskip}
   \label{fig:DatasetCompare}
\end{figure}

\begin{figure}[t]
  \centering
   \includegraphics[width=\linewidth]{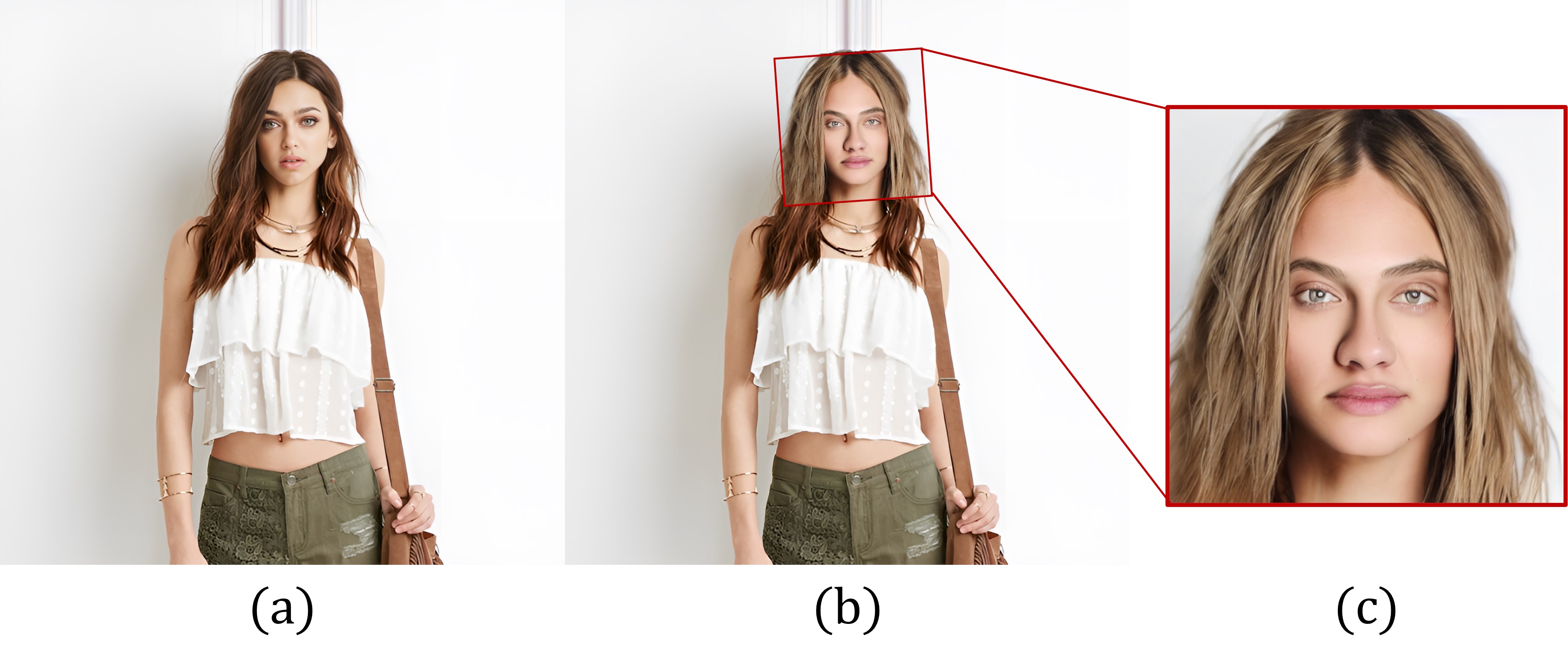}
   \caption{When existing methods that use face-centered cropped datasets~\cite{deepfacelab, reface, faceX} handle cases where the head-swapped image need to be pasted back with the original body image, they could generate inharmonious outcomes because the remaining parts in the body image cannot be addressed. The original body image (a) has brown hair, while the head-swapped image (c) has blonde hair, causing the final image (b) to show a noticeable color inconsistency in the hair. 
    }
   \vspace{-\baselineskip}
   \label{fig:pasteback}
\end{figure}

These challenges stem from the use of face-centered cropped datasets with primarily frontal-facing views and masking methods. As shown in ~\cref{fig:DatasetCompare}(a), most head-swapping methods~\cite{deepfacelab, reface, faceX} perform head swapping only on face-centered cropped data, which may not fully contain the hair. Additionally, the masking methods proposed in~\cite{deepfacelab, reface, faceX} are optimized for these types of images and, thus lacking the ability to create a seamless transition between the head and body. Therefore, if users require head swapping in images that include the full head as well as the body, previous methods need an additional phase to paste the head back into the body, often resulting in inharmonious images. As this issue is clearly shown in  ~\cref{fig:pasteback}, if the original head in the body images has long hair beyond the cropped area, remnants of the hair remain visible. Furthermore, datasets with primarily frontal-facing views lack robustness to diverse facial orientations. Although there have been attempts to address this issue, such as HeSer, it operates in a few-shot manner ~\cite{heser} by utilizing a wide range of view images extracted from video. This highlights the need for a more challenging dataset that includes upper body images with diverse facial orientations and has only a single image per person. To the best of our knowledge, no existing method efficiently addresses these real-world complexities in a zero-shot manner.

Therefore, we design a novel zero-shot head swapping approach, named Head Injection Diffusion (HID),  with proposing Inverse-Orthogonal Mask (IOMask) and hair injection module, which are components of our approach to perform head swapping in real-world scenarios with this challenging dataset. Our approach not only handles diverse facial orientations in a zero-shot setting but also leverages a SHHQ~\cite{shhq} dataset that showcases the entire upper body. This approach combines the strengths of existing datasets while addressing the limitations of previous head swapping methods as illustrated in ~\cref{fig:DatasetCompare}. To tackle the remaining challenges in ensuring that the head of the head image blends naturally onto the body of the body image, we propose our IOMask, which automatically generates a context-aware mask. In head swapping tasks, automating the mask generation process is essential for real-world applications. Aligning the orientation of the head of the head image with head of the body image is crucial, but it is challenging for users to predict and provide such a mask manually and existing automatic masking methods have limitations as aforementioned. We address this issue by introducing the IOMask, extracted within HID, is obtained by leveraging the orthogonal component between the reconstruction condition and the editing condition, both of which originate from the inverted latent.
Furthermore, to preserve face ID, face shape and hairstyle, we build our approach PhotoMaker V2~\cite{pmv2}
\footnote{The code has been released but the paper has not been published yet.}, a state-of-the-art identity-oriented personalization model. Inspired by this method, we introduce our hair injection module to transfer the hair information more precisely.

Our main contributions are as follows:

\begin{itemize}
    \item  We propose a zero-shot head swapping approach, HID, with introducing IOMask and hair injection module, which ensures seamless blending and better preserving intricate hairstyle details, leading to robust performance in real-world applications by using challenging dataset. 
    
    \item 
    We introduce IOMask, a novel approach that automatically generates realistic, context-aware masks for seamless head-body integration. This approach effectively blends the head with the body and removes any remains of the original body image, enhancing the practicality of head swapping for real-world applications.
    
    \item Through experimental results, we demonstrated state-of-the-art head swapping performance in handling complex scenarios.

\end{itemize}

%% file: sec/2_relatedwork.tex
\section{Related work}
\label{sec:relatedwork}

\subsection{Head Swap}

Although there is relatively sparse research on head-swapping tasks, existing studies can be divided into two categories: few-shot approaches~\cite{heser, deepfacelab} and zero-shot approaches~\cite{hsdiffusion, faceX, reface}. Few-shot methods, such as those proposed by ~\cite{heser, deepfacelab}, utilize videos to give a few shots of head input from different views so that the head in the head image is better aligned with the head in the body image.
On the other hand, HSDiffusion~\cite{hsdiffusion} is a zero-shot diffusion-based method but just aligns the center points of the head and body images and cutting and pasting the components accordingly. In other words, it does not align the direction of the head in the head image with the direction of the head in the body image. Therefore, HSDiffusion does not precisely align with traditional head swapping task.
Other zero-shot approaches, like FaceX~\cite{faceX} and REFace~\cite{reface}, also leverage diffusion models for head swapping. \cite{faceX} is a unified method for handling diverse facial tasks and \cite{reface} is a face-swapping method adapted to support head swapping. 

However, most of these existing head swapping methods~\cite{deepfacelab, faceX, reface} has a critical weakness in that they rely on face-centered cropped datasets with primarily frontal-facing view. Since head swapping  is type of an inpainting task, generating an appropriate mask for the edited region is essential, but the mask generation process in ~\cite{deepfacelab, faceX, reface} are optimized to these kinds of datasets. In practical applications, this often results in disharmonious outputs, especially when uncropped areas require adjustments for seamless head integration. Furthermore, even though HeSer~\cite{heser} is not exactly corresponding to this problem, it is a few-shot based approach, which is less efficient than zero-shot. Our proposed approach performs in a zero-shot manner and uses dataset that contains upper body images with diverse facial orientations, enabling more practical and seamless head swapping in real-world applications.

\subsection{Diffusion Models}

Diffusion models~\cite{sd, ho2020denoising, song2020denoising} have achieved impressive advancements in generating images based on text prompts~\cite{imagic, ramesh2022hierarchical, rombach2022high, saharia2022photorealistic} drawing significant attention in recent years. Their exceptional performance is largely due to the availability of high-quality, large-scale text-image datasets~\cite{changpinyo2021conceptual, schuhmann2022laion}, continuous improvements in foundational models~\cite{chen2023pixart, peebles2023scalable}, enhancements in conditioning encoders, and better control mechanisms~\cite{mou2024t2i, li2023gligen, ipadapter, controlnet, ip2p}.

\paragraph{Additional condition.}
ControlNet~\cite{controlnet} and T2I-Adapter~\cite{mou2024t2i} improve diffusion models by encoding spatial information, such as edges, depth, and human pose, to offer more control over the generated content through spatial conditioning alongside text prompts. ControlNet~\cite{controlnet} achieves this by cloning the base model's encoder and incorporating conditioning data as residuals into the model’s hidden states, enhancing coherence with spatial cues. IP-Adapter~\cite{ipadapter} further extends control by using high-level semantics from a reference image, projecting its embedding into the text encoder’s space to enable image generation influenced by both visual and textual prompts.

\paragraph{Identity-oriented Diffusion Model.} 
Identity-oriented diffusion models ~\cite{photomaker, pmv2, arc2face, instantid} enable efficient, high-fidelity identity synthesis. 
PhotoMaker~\cite{photomaker} achieves this by fusing ID embedding with specific text embeddings. PhotoMaker V2~\cite{pmv2} improves upon PhotoMaker by utilizing more data and introducing an improved ID extraction encoder.
InstantID~\cite{instantid} employs a cross-attention-based injection technique similar to IP-Adapter~\cite{ipadapter}, applying it independently to text embeddings. 
Arc2Face~\cite{arc2face} integrates face recognition features from ArcFace~\cite{arcface}, fusing them before passing through the text encoder.

\subsection{Image Editing}
Recently, leveraging the strong generative capabilities of diffusion models, several studies have aimed to facilitate image editing that aligns closely with user intent~\cite{p2p, diffedit, ip2p, watchyoursteps, imagic}. One critical aspect of image editing is local editing, where specific areas of an image are modified based on user guidance. Many methods~\cite{diffedit, watchyoursteps, invedit} address this by employing masks to define the regions requiring adjustment.

In general, masks for local editing within diffusion models can be obtained through two main approaches: one using difference between predicted noises and the other leveraging attention-based~\cite{p2p, attendandexcite} techniques. Using predicted noise's difference methods~\cite{diffedit, invedit, watchyoursteps} use of the disagreement in predictions of stable diffusion~\cite{sd} with source and target captions. This method is advantageous due to their computational efficiency, providing faster editing capabilities. However, they can suffer from inaccuracies, which we aim to mitigate with our proposed approach. On the other hand, attention-based techniques~\cite{p2p,masactrl,attendandexcite,focuson} are typically more precise but have the drawback of slower processing speeds. Although segmentation masks could also be used to delineate editing areas, they come with limitations in certain transformations, such as changing short hair to long hair, where they may lack adaptability. Our method addresses these challenges by improving the accuracy, providing a balanced solution that enhances both precision and speed in local editing tasks.

%% file: sec/3_method.tex
\begin{figure*}
  \centering
   \includegraphics[width=\linewidth]{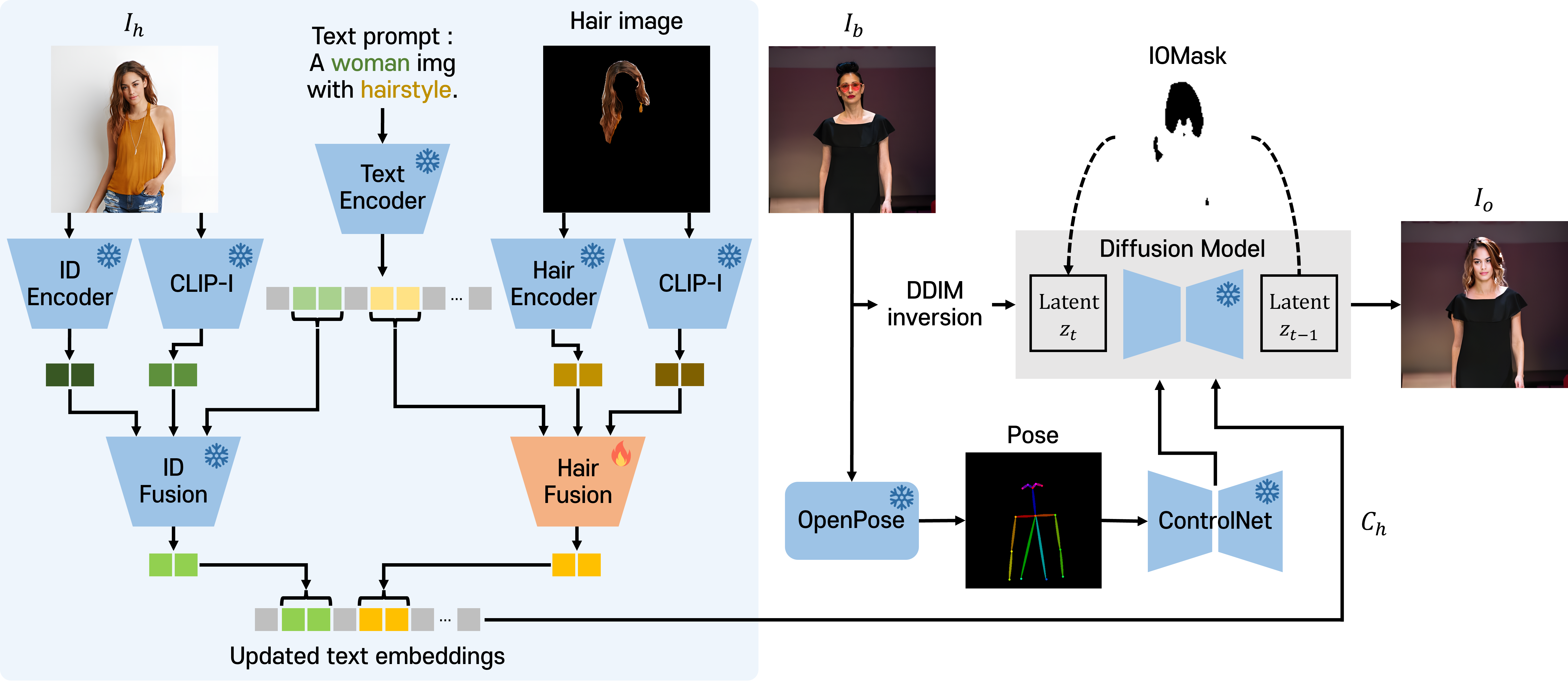}
   \caption{\textbf{Overview of HID.} Our HID consists of two main stages (left and right). In the left stage (blue region), we obtain updated text embeddings by fusing embeddings. 
   These fused ID embeddings and fused hair embeddings replace the part of original text embeddings, resulting in updated text embeddings. In the right stage (white region), the final output $I_o$, a head swapped image, is generated. DDIM inversion is performed to reconstruct the image while leveraging our IOMask to infer which parts of the body image should be removed, thereby generating the head image. During this process, the updated text embeddings obtained in the left stage, along with the output from ControlNet, serves as conditioning inputs for the diffusion model.}
   \vspace{-\baselineskip}
   \label{fig:framework}
\end{figure*}

\section{Method}
We propose a novel approach, HID, that takes a body image \( I_b \) and a head image \( I_h \) as inputs and outputs a swapped image \(I_o\), where the head in \(I_b\) is seamlessly replaced with the one in \(I_h\). We also introduce IOMask for precise mask extraction to guide head integration (Sec.\ref{iomask}) and utilizes hair injection module to generate heads that match the input head images (Sec.\ref{hairinjection}). Together, these components form our cohesive head-swap method (Sec.\ref{inference}). ~\cref{fig:framework} provides an overview of our process.

\subsection{Preliminary}

\textbf{Classifier-free guidance.}
In text-guided generation, a significant challenge is managing the enhanced influence of the conditioned text on the output. To address this, \cite{cfg} introduced a technique known as classifier-free guidance, CFG, which combines an unconditional prediction with a conditioned prediction to achieve the final result.
Formally, let \( \varnothing = \psi("") \) represent the embedding of an empty or null text, and let \( w \) denote the guidance scale parameter. The CFG prediction is then formulated as follows:
\begin{equation}\label{eq1}
{\epsilon}_{\theta}(z_t, t, C, \varnothing) =  \epsilon_{\theta}(z_t, t, \varnothing)
+ w \cdot (\epsilon_{\theta}(z_t, t, C) -  \epsilon_{\theta}(z_t, t, \varnothing)).
\end{equation}

\vspace{-10pt}
\paragraph{DDIM inversion.}
A simple inversion method has been suggested for DDIM sampling \cite{ddim}, which relies on the premise that the ordinary differential equation process can be retraced in reverse if the step size is sufficiently small.
\begin{equation}
z_{t+1} = \sqrt{\frac{\alpha_{t+1}}{\alpha_t}} z_t  + \left( \sqrt{\frac{1}{\alpha_{t+1}} - 1} - \sqrt{\frac{1}{\alpha_t} - 1} \right) \cdot \epsilon_{\theta}(z_t, t, C).
\end{equation}
Put differently, the diffusion process follows a reversed trajectory, going from \(z_0\) to \(z_t\) rather than from \(z_t\) to \(z_0\), with \(z_0\) initialized as the encoding of a real input image. Moreover, DDIM Inversion uses CFG with \(w = 1\) that means text-conditioned predicted noise $\epsilon_{\theta}(z_t, t, C)$ is used.

\vspace{-10pt}
\paragraph{PhotoMaker V2.}
Given a few ID images to be customized, PhotoMaker V2~\cite{pmv2} generates a new photo-realistic human image that retains the characteristics of the input IDs. This capability is achieved by updating text embeddings. Specifically, the text embedding for a particular class is fused with the image embeddings extracted by CLIP~\cite{clip} image encoder and the ID embeddings extracted by ID encoder~\cite{insightface}.
In this paper, we refer the model that creates the fused embedding, as ID Fusion model. The output of ID Fusion model replaces the class text embedding before it is input into the U-Net.

\subsection{IOMask}
\label{iomask}
For accurate head-swapping, obtaining a mask for head region is crucial to generate on the body image \(I_b\). Inspired by~\cite{diffedit, watchyoursteps, invedit}, we achieve this by extracting the mask within our model.
Given an body image \(I_b\), we first apply DDIM~\cite{ddim} Inversion with our model, obtaining the latent \(\hat{z}_t\) at a specific time step \(t\).
If we use DDIM Inversion with guidance scale \(w = 1\), that provides a faithful reconstruction when adapting the denoising guidance keep the scale same.
For specific time-step \(t\), we can get $\epsilon_{\theta}(\hat.{z}_t, t, C_b)$ from denoising \(\hat{z}_t\), where \(C_b\) means body image's text embeddings, and established this for a baseline.
The Inverse-Orthognal map, called IO map, is then determined by the areas that differ under head-conditioned predicted noise ${\epsilon_{\theta}}(\hat{z}_t, t, C_h, \varnothing)$, where \(C_h\) means head image's text embeddings with fused ID embeddings and fused hair embeddings.
Note that CFG is conducted for head-conditioned predicted noise.
For simplicity, we'll omit notation \(\hat{z}_t\) and \(t\).

\begin{figure}[t]
  \centering
   \includegraphics[width=0.5\linewidth]{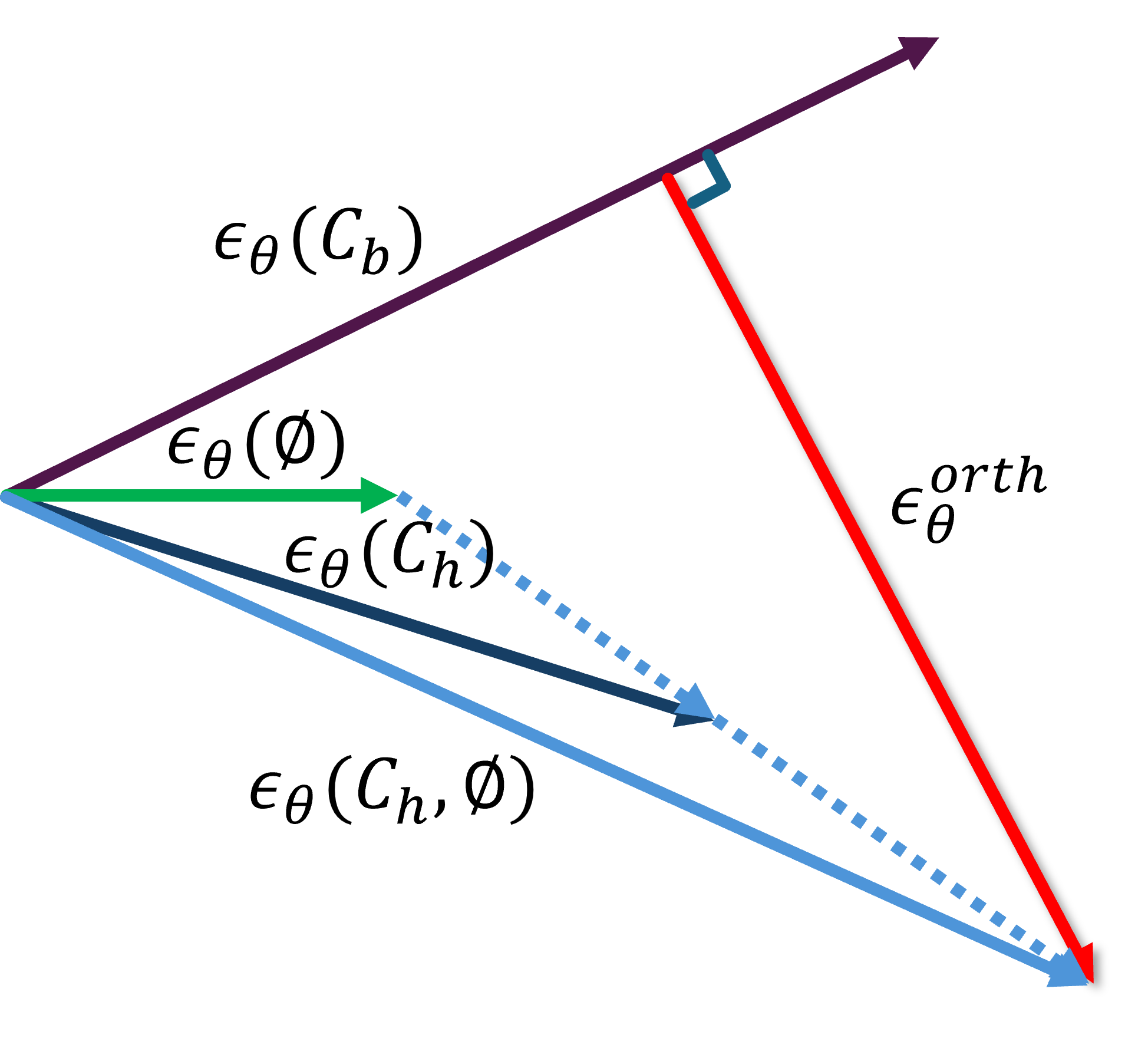}
   \caption{Visualization of obtaining $\epsilon_{\theta}^{orth}$ in vector form.}
   \label{fig:IOMap}
   \vspace{-\baselineskip}
\end{figure}

To obtain IO map, we compare $\epsilon_{\theta}(C_h, \varnothing)$ and $\epsilon_{\theta}(C_b)$. Using $\epsilon_{\theta}(C_b)$ as the reference, we compute the mask by isolating vector components of $\epsilon_{\theta}(C_h, \varnothing)$ orthogonal to the body condition, rather than relying on absolute differences $\epsilon_{\theta}(C_h, \varnothing)$ - $\epsilon_{\theta}(C_b)$. We provide a visualization of this process in vector form in Fig.~\ref{fig:IOMap}.
When adjusting normalization and thresholding, the distribution of value magnitudes is critical. Values aligned in the similar direction should be smaller, while those in differing directions should be larger, as the head area will be oriented differently and should be emphasized.
Therefore, we base our calculations on the components orthogonal to the reference $\epsilon_{\theta}(C_b)$. Thus, the formulation is given by:
\begin{equation}
    \epsilon_{\theta}^{orth} =  \epsilon_{\theta}(C_h, \varnothing) - \frac{<\epsilon_{\theta}(C_b),  \epsilon_{\theta}(C_h, \varnothing)>}{\|\epsilon_{\theta}(C_b)\|^2} \epsilon_{\theta}(C_b)
\end{equation}
After normalizing the IO map between 0 and 1, we observe the presence of dotted artifacts. To address this issue and improve robustness, we apply a Gaussian filter. The Inverse-Orthogonal Mask, called IOMask, \( \mathcal{M} \) is then computed as follows:
\begin{equation}
    \mathcal{M} = \mathbf{1}_{(GF(|\epsilon_{\theta}^{orth}|) \geq \tau )},
\end{equation}
where \( GF \) denotes the Gaussian filter function, and \( \tau \in [0, 1] \) is the threshold parameter.

\subsection{Hair Injection Module}
\label{hairinjection}
While PhotoMaker V2~\cite{pmv2} can generate human images that maintain a given identity to varying degrees, it cannot be directly applied to the head-swapping task because it is fundamentally an image generation model, not an image editing model. To address this, we first introduce the IOMask to provide the model with information about the region that needs to be edited.

A remaining challenge is to generate hairstyles that accurately reflect the hairstyle features in $I_h$. Since PhotoMaker V2 relies solely on text prompts for conditioning, aside from ID, we inject hairstyle information from images specifically for the head-swapping task by training the Hair Fusion model inspired by PhotoMaker V2. We train the Hair Fusion model with text prompts that includes 'A man/woman \textit{img} with \textit{hairstyle}.' \textit{img} serves as a trigger word for ID injection, while the embeddings of \textit{hairstyle} is fused with the hair embeddings extracted from the pretrained Hair Encoder~\cite{hairenc} and the hair image embeddings extracted from the CLIP image encoder. The Hair Fusion model comprises a Q-former and MLPs, and the text embeddings are updated such that the fused ID embeddings and fused hair embeddings replace the corresponding original text embeddings. This allows the model to learn a distribution that enables these updated embeddings to condition the UNet, facilitating effective head-swapping. Also, during hair injection module training, SCHP~\cite{schp} was used to extract a mask for regions excluding the background, and a masked loss was applied to reconstruct the person, allowing for effective hairstyle integration.

\subsection{Head Injection Diffusion Model}
\label{inference}
The Head Injection Diffusion model (HID) performs head swapping by combining the hair injection module with IOMask. 
Given a body image \( I_b \) and a head image \( I_h \), the method outputs a swapped image \( I_o \).
To achieve this, $I_h$ is processed through the ID injection module and a hair image is processed through hair injection module. Each module outputs a fused embedding which replaces the corresponding text embeddings to generate the head condition \( C_h \), used as input through cross-attention. Additionally, the body pose is provided to the model via open-pose ControlNet~\cite{controlnet} to ensure alignment.

The denoising process begins with the inversion latent \( \hat{z}_T \), stored during IOMask generation. Starting with this latent ensures that the masked region does not begin without body details, thereby preventing mismatches in skin tones or clothing artifacts. This inversion latent is continuously blended with the mask during updates, promoting a more natural result. Each denoising stage refines a noisy latent \( z_t \) to obtain \( z_{t-1} \).

During each denoising step, given the head condition \( C_h \), the latent \(z_t\) is denoised to latent \( \tilde{z}_t \) and updated as follows:
\begin{equation}
    z_{t-1} = \tilde{z}_{t-1} \odot \mathcal{M} + \hat{z}_{t-1} \odot \left( 1 - \mathcal{M} \right)
    \label{eq5}
\end{equation}
By replacing the unmasked pixels with the inversion latent of the input image, we prevent the generation process from altering any pixels outside the mask. After iterative denoising, the head-swapped image \(I_o\) is obtained.

%% file: sec/4_experiments.tex
\section{Experiments}

\begin{figure*}[t]
  \centering
   \includegraphics[width=1.0\linewidth]{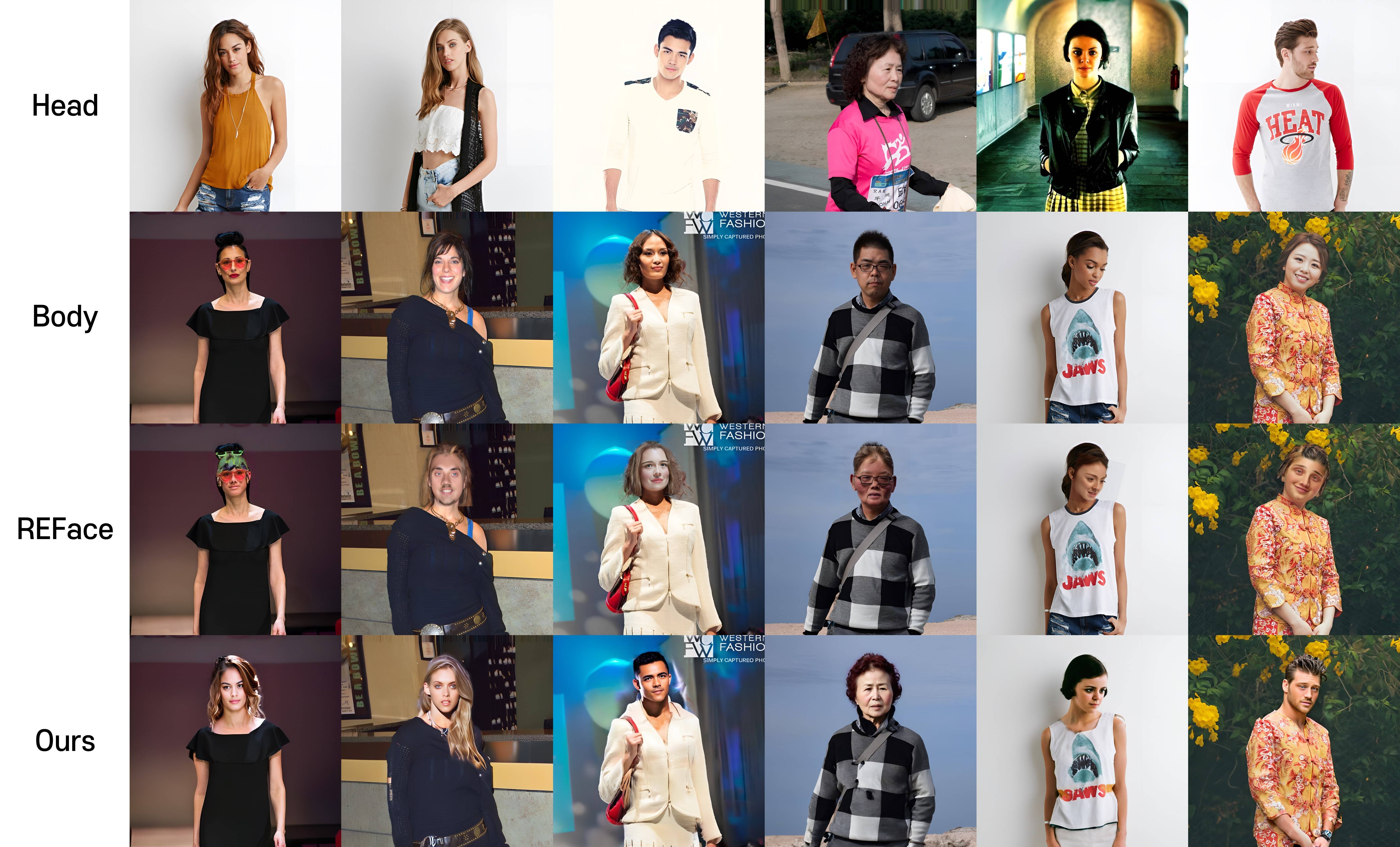}
   \caption{\textbf{Qualitative comparison.} 
    The images in the \textit{Head} row are combined with those in the \textit{Body} row. The last two rows are the head-swapped results produced by each method.}
   \label{fig:qual_result}
  \vspace{-\baselineskip}
\end{figure*}

\subsection{Experimental Setup}
\label{sec:exp}
\paragraph{Dataset.} We use the SHHQ-1.0 dataset~\cite{shhq}, consisting of 40K images, for training and evaluation. The original resolution of the dataset is 1024 $\times$ 512; however, to focus on the head swapping task, we preprocess the images to 512 $\times$ 512 by cropping out the lower part of the body. Given that the maximum resolution that SDXL~\cite{podell2023sdxl} can generate is 1024 $\times$ 1024, we upsample the images to this resolution using GFPGAN v1.4~\cite{gfpgan}. After image preprocessing, we obtain a total of 39,804 images, which are split in a 9:1 ratio. This results in 35,823 images for training and 3,981 images for evaluation. In addition, We generate image captions for training using the state-of-the-art image captioning model, LLaVa-NEXT~\cite{llavanext}, which is based on LLama-3-8B~\cite{dubey2024llama}. To incorporate ID and hairstyle information, we modify the captions to include the phrase 'A man/woman img with hairstyle.'

\vspace{-8pt}
\paragraph{Baselines.}
To the best of our knowledge, there are no existing baselines that align precisely with our head swapping task, which operates in a zero-shot manner, reenacts the head in the head image to match the orientation of the head in the body image, and does not require face-centered cropping or alignment. Therefore, we compare our performance solely with 
REFace~\cite{reface}, the only head swapping method that has officially released its code between zero-shot head swapping method, to demonstrate that our approach outperforms an existing method. 

\vspace{-8pt}
\paragraph{Evaluation.} All images in the evaluation dataset are used as head images, which are randomly paired with other images from the same dataset to serve as body images. We use 'A man/woman image with hairstyle' as a text prompt for the evaluation. The evaluation is conducted using the following metrics: Learned Perceptual Image Patch Similarity (LPIPS)\cite{lpips}, CLIP-I~\cite{clip} and Frechet Inception Distance (FID)\cite{fid}.
To demonstrate the performance of our approach, we calculate LPIPS and CLIP-I using images with masked hair or masked head regions, where the head mask includes both hair and face. We first generate human parsing masks using SCHP~\cite{schp} and use these masks to isolate hair or head regions. After that, we measure LPIPS and CLIP-I between the hair and head regions of the generated images and the corresponding regions in the real images from the evaluation dataset. This allows us to more closely compare the generated image with specific features of the real image to make an assessment of how similar and consistent the targeted areas are. Additionally, we use FID to assess the quality of generated images. However, since REFace~\cite{reface} handles solely on face-centered cropped regions, a fair comparison is difficult. To address this, we crop the same region from our generated images to match the face-centered area used by REFace.

\subsection{Results}

\paragraph{Quantitative results.}
As shown in ~\cref{tab:quan_result}, our HID outperforms REFace~\cite{reface} in both LPIPS and CLIP-I metrics for the head and hair regions. This result indicates that HID more accurately preserves ID and hairstyle in head images, which is essential for the head-swapping task. Moreover, we outperformed REFace on the FID score. This demonstrates that our generated images have superior quality, achieving a more natural integration of head and body features compared to existing methods.

\vspace{-12pt}
\paragraph{Qualitative results.} 
As shown in \cref{fig:headswapDef}, the swapped head must preserve key attributes of the head image, including skin tone, hairstyle, identity, and face shape. \cref{fig:teaser} illustrates how our method effectively achieves these requirements, performing the head swap task while maintaining these essential characteristics. Additionally, thanks to the IOMask, we can generate appropriate masks that allow the creation of long hair where needed, ensuring that only the relevant head region is modified while leaving other areas unchanged.
In \cref{fig:qual_result}, we demonstrate that our method outperforms REFace~\cite{reface}, particularly in challenging cases where head orientation and gender differ significantly. Unlike REFace, which suffers from limitations inherent to the cropping approach—such as hair being cut off and not blending well with surrounding areas—our approach overcomes these issues, maintaining a seamless and coherent appearance even under these challenging conditions. Furthermore, unlike the mask used in REFace, our proposed IOMask effectively removes only the necessary face and hair areas from the body image, enabling a smooth and natural swap.

\subsection{Ablation Study}
\subsubsection{IOMask}
\begin{figure}[t]
  \centering
   \includegraphics[width=1.0\linewidth]{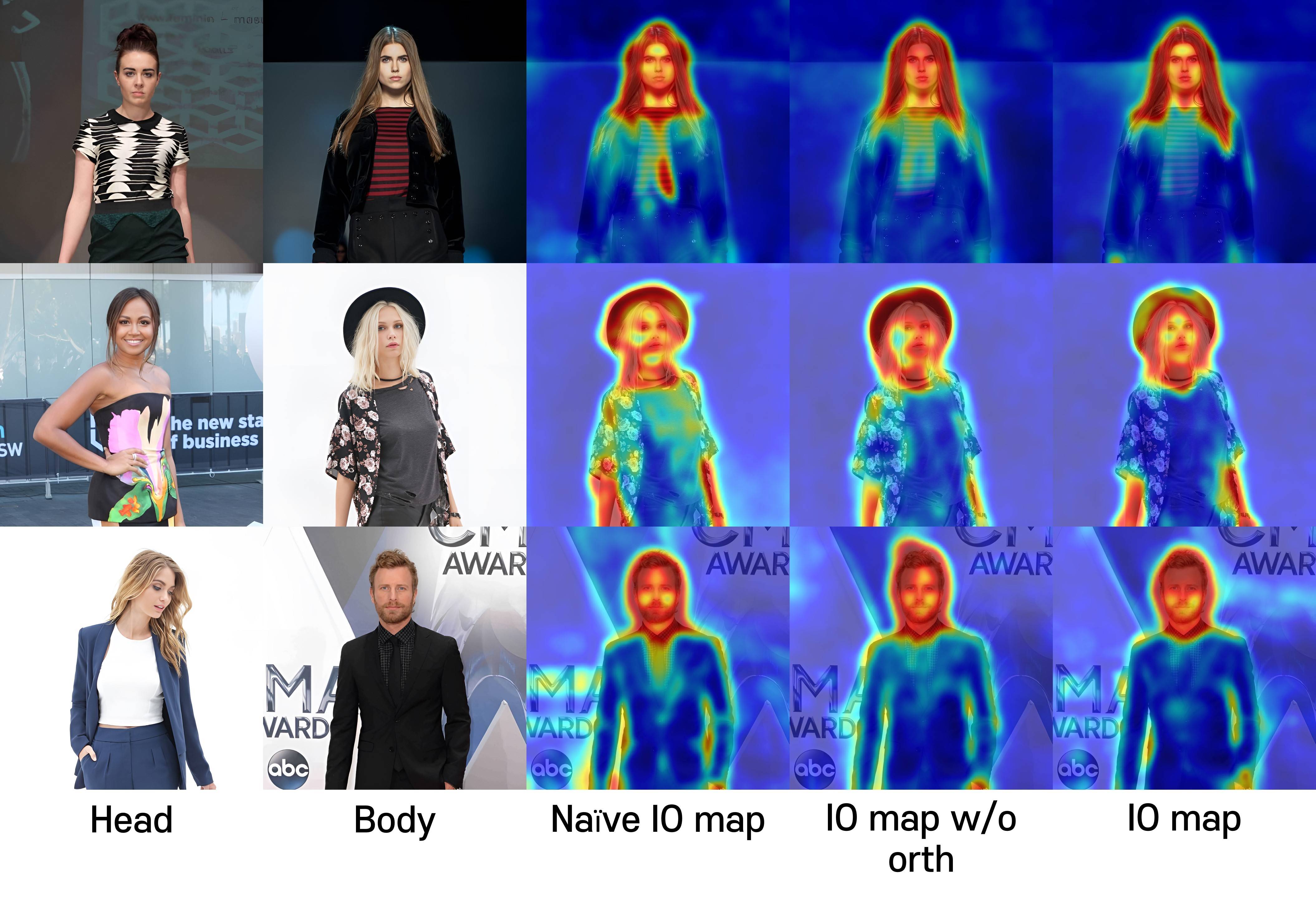}
   \caption{~\textbf{Ablation study on IOMask configurations.} We compare three variants: the naive IO map, the IO map without orthogonal filtering, and the full IO map. The full IO map demonstrates improved focus and precision in identifying relevant areas for head swapping, as indicated by regions of high values in red.} 
   \vspace{-\baselineskip}   
   \label{fig:ablation_iomask}
\end{figure}

We demonstrate the effectiveness of our IOMask through an ablation study, comparing the results of three configurations: the naive IO map, IO map without orthogonal components, and the full IO map.
The naive IO map is defined as $\epsilon_{\theta} = \epsilon_{\theta}(C_h, \varnothing) - \epsilon_{\theta}(C_b, \varnothing)$, where the ID embeddings of both head and body conditions are without further refinement. 
The IO map without orthogonal components is defined as $\epsilon_{\theta} = \epsilon_{\theta}(C_h, \varnothing) - \epsilon_{\theta}(C_b)$
, omitting the orthogonal filtering step present in our full IO map.
We visualized the IO map by overlaying it on body images, with red indicating regions of high values.
As illustrated in ~\cref{fig:ablation_iomask}, the naive IO map introduces substantial randomness and often covers irrelevant areas. In contrast, both the IO map without orthogonal components and the full IO map provide more relevant coverage. Notably, the full IO map focuses more precisely on the regions that have to be swapped.

\subsubsection{HID}

\begin{figure}[t]
  \centering
   \includegraphics[width=1.0\linewidth]{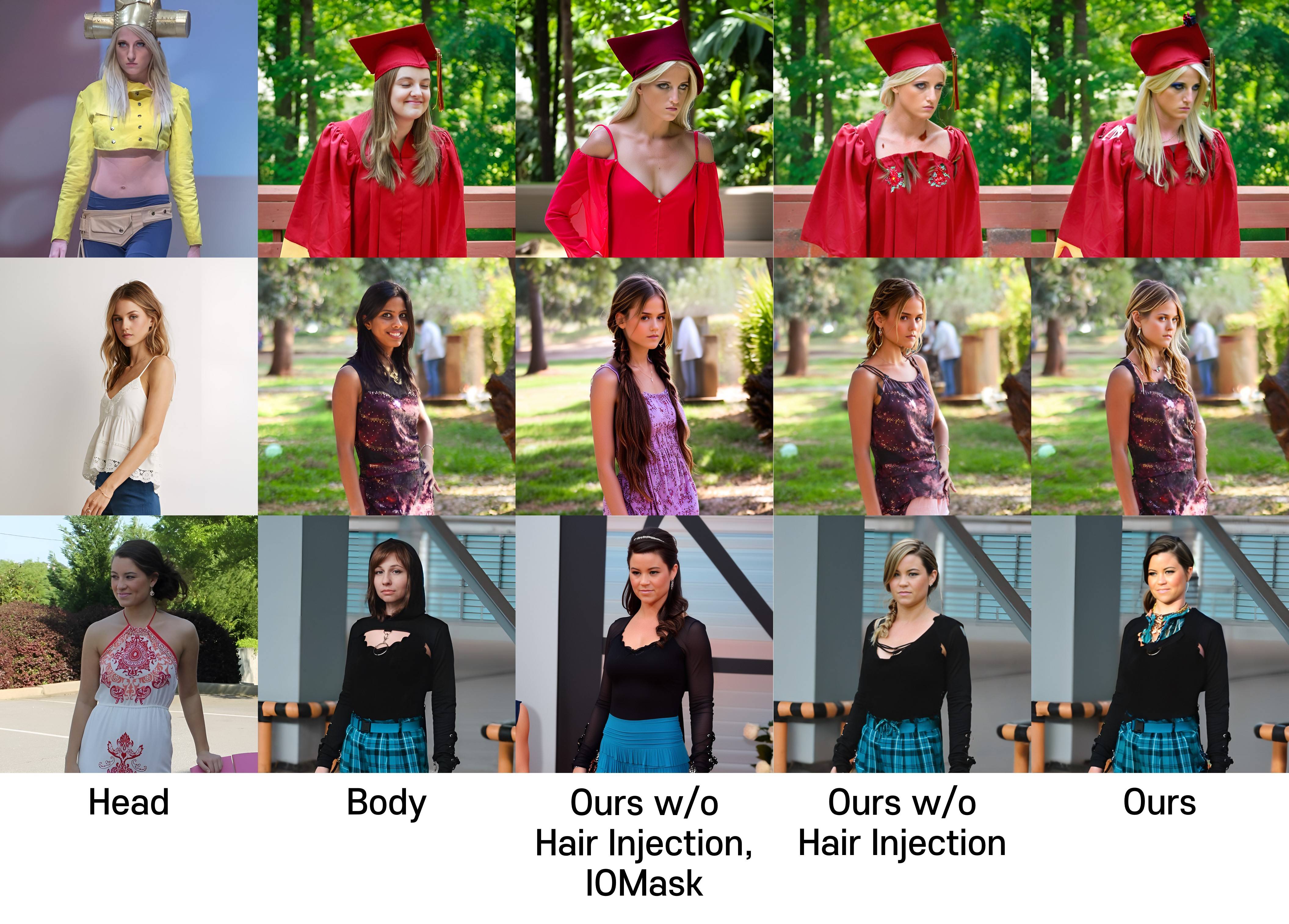}
   \caption{\textbf{Ablation study on components of our approach.} We conducted ablation experiments by removing each component of our method individually. Without the hair injection module, the model fails to retain the head image's hairstyle, as seen in the first row where the hair appears short. Additionally, removing the IO mask results in a complete loss of body image information.}
  \vspace{-\baselineskip}
   \label{fig:ablation_hid}
\end{figure}

We evaluate the performance impact of each element in the HID to show the effect of our methods as shown in \cref{fig:ablation_hid}. So, we sequentially remove the hair injection module and IO Mask, while applying both model with ControlNet and same input text, 'A man/woman img with hairstyle.'

Starting with the our full approach, which includes all components, we achieved optimal results with high-quality head swaps, including detailed hairstyle transfer and effective integration of the head with the body. When the hair injection module was removed, the generated results showed a noticeable reduction in hairstyle detail, as the method could no longer fully transfer the hairstyle from the head image onto the body image. This is because, while PhotoMaker V2 effectively maintains faceID, it does not inherently ensure hairstyle preservation.

Finally, although generation began from the inversion latent, removing the IO Mask resulted in retaining only a very slight amount of the body image and losing its ability to maintain regions of the body image outside the head. This setup led to inconsistencies in other areas of the body image, underscoring the importance of each component for achieving a natural and coherent final output.

\begin{table}[t]
\centering
\vspace{5pt}
\resizebox{\linewidth}{!}{
  \begin{tabular}{@{}l|c|cc|cc@{}}
    \toprule
    \multirow{2}{*}{} & \multirow{2}{*}{FID$\downarrow$} & \multicolumn{2}{c|}{Head} & \multicolumn{2}{c}{Hair} \\   
                      &        & LPIPS$\downarrow$ & CLIP-I $\uparrow$ & LPIPS$\downarrow$ & CLIP-I $\uparrow$ \\
    \midrule
    REFace            &  40.7162  &    0.0770           &     0.7867       &   0.0658         &     0.8563            \\
    Ours              &  \textbf{37.1879} &   \textbf{0.0721}  &  \textbf{0.8512} &  \textbf{0.0596} &     \textbf{0.8686}     \\
    \bottomrule
  \end{tabular}
}
  \caption{\textbf{Quantitative comparison.} Ours outperforms REFace~\cite{reface} across all metrics.}
  \label{tab:quan_result}
   \vspace{-\baselineskip}
\end{table}

%% file: sec/5_conclusion.tex
\section{Conclusion}
We propose the HID approach, an efficient head swapping method aimed at real-world applications. Our approach allows the head image to blend seamlessly with the body image while preserving original details, resulting in a natural outcome. Despite these significant improvements over previous approaches, there are still areas that require refinement. For example, regions beyond the head, like the hands, neck, or other skin areas, may experience unintended changes. In the \cref{fig:teaser}, the second example from the right shows a case where the hands were masked to adjust their color, allowing them to match the skin tone of the head image. However, while the skin tone was transferred successfully, the pose and fine details of the original body image's hands were not preserved, which highlights a limitation of our approach. Unlike previous methods that rely heavily on cropping or head-centric masking techniques, our approach takes into account which areas should be edited when conditioned with the head image, allowing for more flexibility and potential for improvement. Nevertheless, our method shows potential for further advancements, and future work could focus on better preserving details of these regions while still achieving seamless skin tone transfer.

%% file: sec/Acknowledgment.tex
\section*{Acknowledgement}
This work was supported by Institute for Information \& communications Technology Planning \& Evaluation(IITP) grant funded by the Korea government(MSIT) (RS-2019-II190075, Artificial Intelligence Graduate School Program(KAIST)). This work was supported by the National Research Foundation of Korea(NRF) grant funded by the Korea government(MSIT) (No. RS-2025-00555621). We appreciate the high-performance GPU computing support of HPC-AI Open Infrastructure via GIST SCENT.

%% file: sec/X_suppl.tex
\clearpage
\setcounter{page}{1}
\maketitlesupplementary

\begin{figure}[t]
  \centering
  \includegraphics[width=1.0\linewidth]{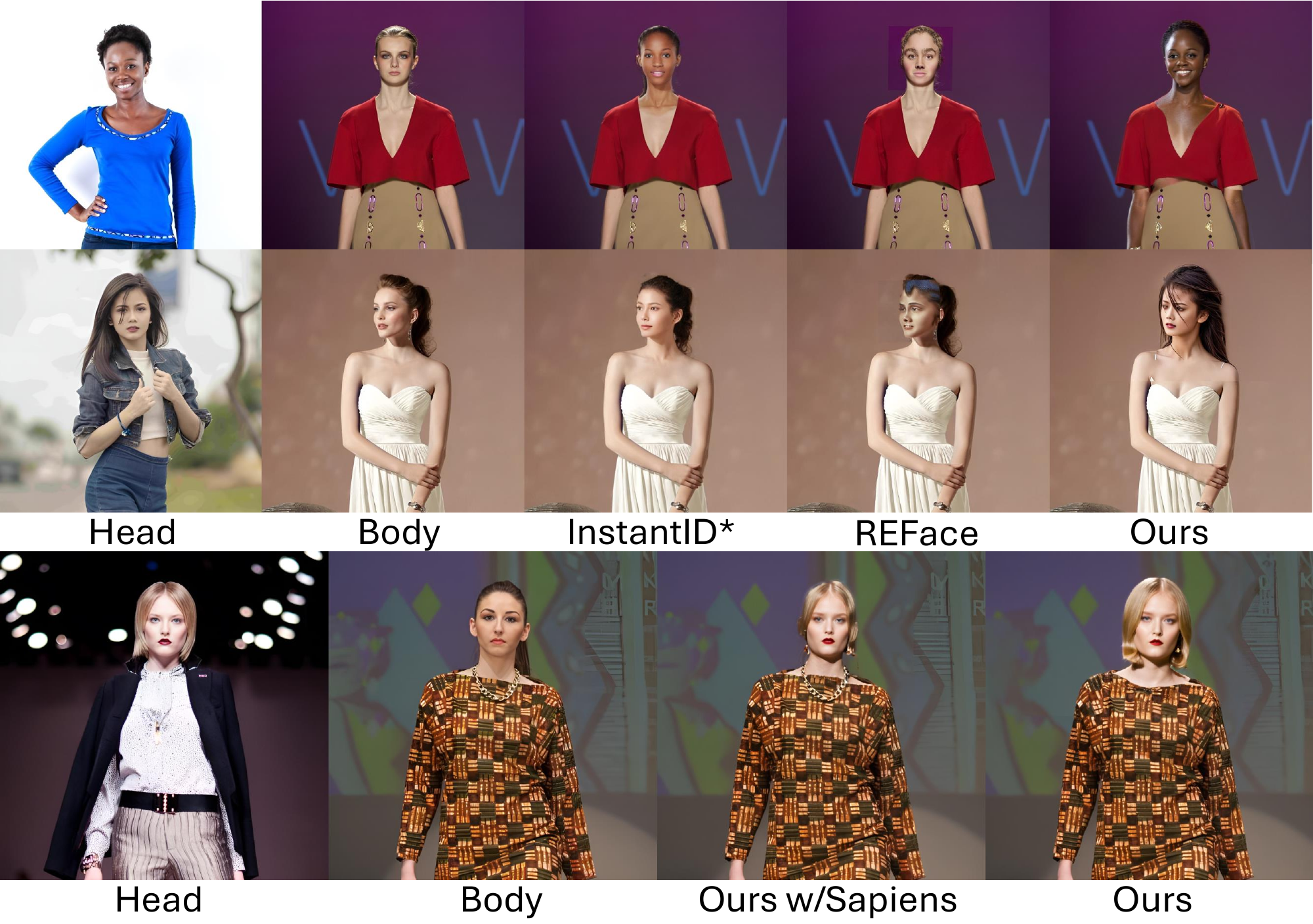}
   \caption{Additional qualitative comparisons between InstantID and Ours, as well as Ours with Sapiens mask and Ours.}
   \label{fig:onecol}
   \vspace{-10pt}
\end{figure}

\section{Implementation Detail}
\subsection{Training}
Training is implemented using the HuggingFace diffusers library~\cite{diffusers} with data type casting to bfloat16 to enhance training efficiency. Additionally, it is conducted on 8 A100 40GB GPUs over 210,000 steps, using a batch size of 2 per GPU. To fuse the embeddings generated by the pretrained hair encoder, which are in the shape [18, 512], a 1$\times$1 convolutional layer is employed to reshape them to [1, 512]. To train the Hair Fusion model and convolutional layer, the AdamW optimizer~\cite{adamw} is used with a constant learning rate of 0.00001 and a weight decay of 0.01.

\subsection{IOMask} \label{implementation}
IOMask is used to extract precise masks for head generation. We set the threshold $\tau = 0.6$ and perform 40 denoising steps out of 50, corresponding to 80\% noise levels. This approach ensures robust mask generation tailored to the head-swap task without requiring preprocessing steps like alignment or cropping.

\section{Problem Definition}
Face swapping can be defined in various ways, such as replacing only the eyes, nose, and mouth while preserving the original face shape. In our case, we define a “head swap” as the task of swapping only the head, which is considered part of the human body. Consequently, long beards should be reflected in the swap, whereas accessories should not. To ensure that accessories are not swapped, we refined the training dataset accordingly. As shown in Fig.7 in manuscript, facial hair is partially preserved in the swap to a reasonable extent. Moreover, for a seamless transition, it is necessary to process not only skin of the neck area, which connects to the head, but also the hands. Our IOMask has the potential to enable this process as shown in Fig.~\ref{fig:onecol}, first row.

\begin{table}[t]
\vspace{3pt}
\centering
\resizebox{\linewidth}{!}{
  \begin{tabular}{@{}l|cc|cc|c|c|c@{}}
    \toprule
    \multirow{2}{*}{} & \multicolumn{2}{c|}{Head (crop)} & \multicolumn{2}{c|}{Hair (crop)} & \multirow{2}{*}{ID sim $\uparrow$} & \multirow{2}{*}{Head Pose $\downarrow$} & \multirow{2}{*}{Recon. $\uparrow$} \\ 
    & LPIPS$\downarrow$ & CLIP-I $\uparrow$ & LPIPS$\downarrow$ & CLIP-I $\uparrow$ \\
    \midrule
    REFace&0.4932&0.7430&0.3189&  0.8312  &  0.5197  &  \textbf{20.33} & \textbf{0.6397}  \\
    InstantID* &0.4802&0.8090&0.3251& 0.8470 & 0.2821 &  20.64 & 0.6308 \\
    Ours &\textbf{0.4567}&\textbf{0.8615}& \textbf{0.2966} &  \textbf{0.8644}  & \textbf{0.5546} &  21.66 & 0.6286\\
    \bottomrule
  \end{tabular}
    }
  \caption{Quantitative comparison.}
  \label{tab:quan_result}
   \vspace{-18pt}
\end{table}

\section{Analysis on Qualitative Results} 
\label{analysis}
The reason parts other than the head are sometimes altered, called hallucinations or artifacts, is that the mask leaves those areas uncovered. As mentioned in the Sec.1, if we want to generate a head independent of the original body’s head, using predicted noise’s difference methods are effective. However, these methods have a drawback: they often open up areas that do not need to be changed. IOMask helps minimize these issues while still preserving the advantages of such approaches as shown in Fig.~\ref{fig:onecol}, second and third row.
Note that REFace is a method that uses the crop-and-paste-back approach, ensuring that other parts of the image remain unchanged which has its own limitations discussed in Sec.1.
When using an off-the-shelf mask, it only fills the head region of $I_b$, which can lead to issues such as the inability to generate long hair or the appearance of sharp skin boundaries, as shown in Fig.\ref{fig:onecol}. In contrast, IOMask allows for more flexible and natural generation.

We use a dataset that includes the entire upper body. Therefore, rather than providing only head orientation information, we utilize OpenPose, which offers richer information to ensure better alignment with the body. It is also worth noting that, as shown in Fig.5, openpose has difficulty capturing fine details of hands~\cite{fu2024adaptive}. Since our focus is on head swapping, this limitation lies outside the scope of our work. However, it is an important aspect to consider for future improvements, particularly in full-body swap scenarios.

\section{Additional Quantitative Comparisons}
We additionally evaluated our method using four metrics: ID similarity with the face recognition model~\cite{arcface}  head pose L2 distance with the head pose estimation network, HopeNet~\cite{Ruiz_2018_CVPR_Workshops}, LIPIS and CLIP-I for cropped images to compare with REFace, as well as a reconstruction metric, MSE, to assess the degree of preservation outside of the head area. Furthermore, we conducted an additional comparison with SDXL Inpainting InstantID~\cite{instantid}, using Sapiens~\cite{khirodkar2024sapiens} mask, with openpose ControlNet. InstantID not only struggles to preserve identity compared to our method but also fails to transfer hair.
Our method outperformed in all aspects except for head pose and reconstruction. The weaknesses observed in head pose and reconstruction can be attributed to the trade-off between flexibility and preservation, which arises from the different masking methods.

\section{Additional Qualitative Results}
We provide additional qualitative results generated by our proposed approach.

\begin{figure*}[t]
  \centering
   \includegraphics[width=0.9\linewidth]{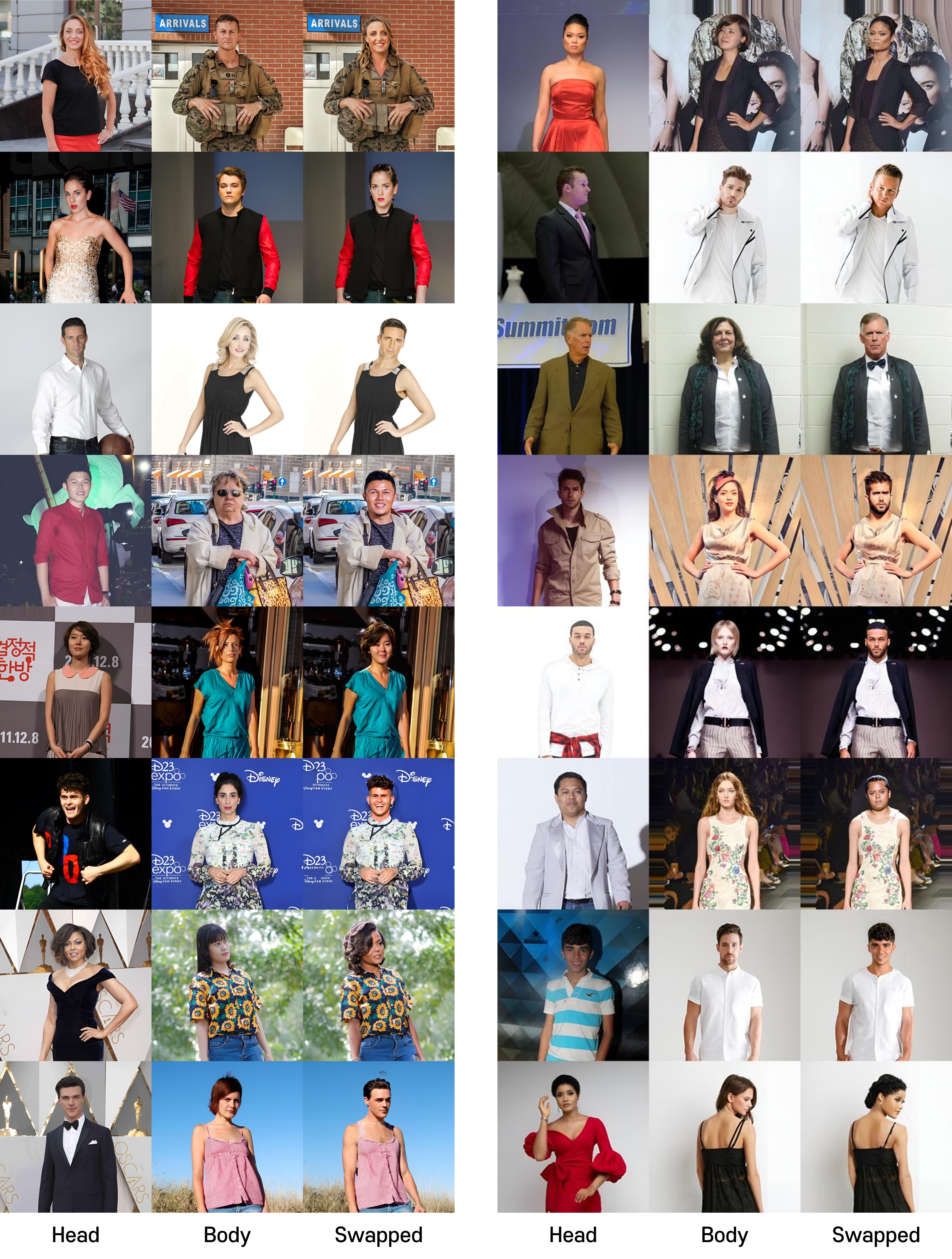}
   \caption{\textbf{Additional qualitative results.} The head in the images of \textit{Head} column is seamlessly combined with the body in the images of \textit{Body} column by the proposed method, HID, resulting in head-swapped images in the \textit{Swapped} column.}
   \label{fig:additional_qual}
\end{figure*}